\documentclass[letterpaper]{article} 
\usepackage{aaai2026}  
\usepackage{times}  
\usepackage{helvet}  
\usepackage{courier}  
\usepackage[hyphens]{url}  
\usepackage{graphicx} 
\usepackage{natbib}  
\usepackage{caption} 
\usepackage{algorithm}
\usepackage{algorithmic}

\usepackage{newfloat}
\usepackage{listings}
\usepackage{array, booktabs, colortbl,multirow,adjustbox,float}
\usepackage{amsmath}
\usepackage{amssymb}
\usepackage{amsfonts}
\usepackage{pifont}
\usepackage{xspace}
\usepackage{subcaption}
\usepackage{xcolor}

\newcommand{\ie}{\textit{i.e.}\xspace}
\newcommand{\eg}{\textit{e.g.}\xspace}
\newcommand{\etal}{\textit{et al.}\xspace}
\newcommand{\tool}{ExpertAD\xspace}

\DeclareCaptionStyle{ruled}{labelfont=normalfont,labelsep=colon,strut=off} 
\floatstyle{ruled}
\newfloat{listing}{tb}{lst}{}
\floatname{listing}{Listing}
\title{\tool: Enhancing Autonomous Driving Systems with Mixture of Experts}
\author{
    Haowen Jiang, 
    Xinyu Huang, 
    You Lu,
    Dingji Wang, 
    Yuheng Cao, 
    Chaofeng Sha, 
    Bihuan Chen\thanks{Corresponding Author},  
    Keyu Chen, 
    Xin Peng
}
\affiliations{
    College of Computer Science and Artificial Intelligence, Fudan University, Shanghai, China\\

}

\usepackage{bibentry}

\begin{document}

\maketitle

\begin{abstract}
    Recent advancements in end-to-end autonomous driving systems (ADSs) underscore their potential for perception and planning capabilities. However, challenges remain. 
    Complex driving scenarios contain rich semantic information, yet ambiguous or noisy semantics can compromise decision reliability, while interference between multiple driving tasks may hinder optimal planning. Furthermore, prolonged inference latency slows decision-making, increasing the risk of unsafe driving behaviors.
    To address these challenges, we propose \tool, a novel framework that enhances the performance of ADS with Mixture~of~Experts~(MoE) architecture. 
    We introduce a Perception Adapter (PA) to amplify task-critical features, ensuring contextually relevant scene understanding, and a Mixture of Sparse Experts (MoSE) to minimize task interference during prediction, allowing for effective and efficient planning.
    Our experiments show that \tool reduces average collision rates by up to 20\% and inference latency by 25\% compared to prior methods. We further evaluate its multi-skill planning capabilities in rare scenarios (e.g., accidents, yielding to emergency vehicles) and demonstrate strong generalization to unseen urban environments. Additionally, we present a case study that illustrates its decision-making process in complex driving scenarios.
    \end{abstract}  


\section{Introduction}\label{sec:intro}

End-to-end autonomous driving systems (ADSs) are gaining growing attention, supported by advances in computational power that enable efficient data processing and real-time decisions. Most vision-based ADSs~\cite{hu2022st,jiang2023vad} use multi-view camera inputs to produce 2D representations for downstream perception, prediction, and planning.
Recent efforts integrate large language models (LLMs)\cite{renz2024carllava,shao2024lmdrive,xu2024drivegpt4} and world models (WMs)\cite{wang2025diffad, wang2024driving,wang2023drivedreamer}, leveraging their generalization and predictive capabilities.
However, most deployed systems~\cite{tian2024drivevlm,chen2025solve} still rely on traditional end-to-end ADSs to generate final trajectories. Even with LLMs or WMs, end-to-end ADSs are typically retained as fallbacks to ensure reliable execution. In this work, we aim to enhance end-to-end ADSs by improving planning effectiveness and reducing inference latency.

\begin{figure}[!t]
    \centering
    \includegraphics[width=0.9\linewidth]{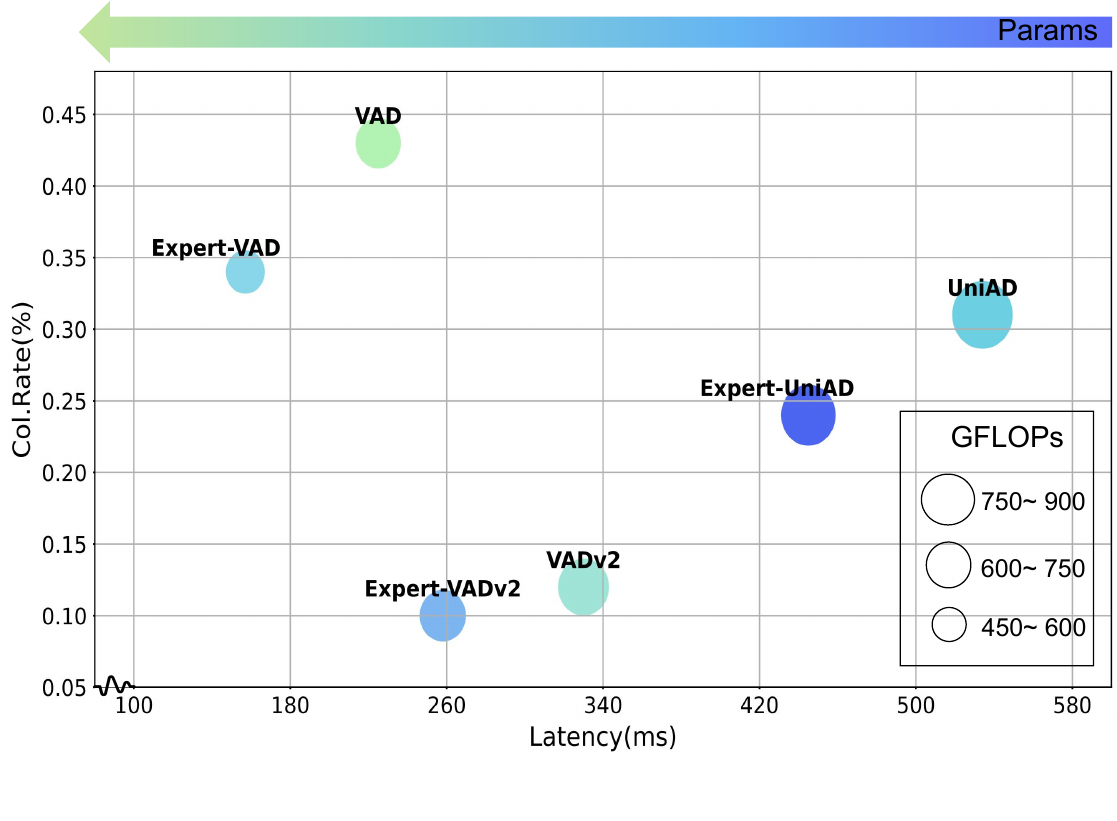}
    \caption{\textbf{Collision rate and latency trade-offs across different models.} 
    \tool exhibits~substantial improvements in planning effectiveness while reducing inference latency, as measured on NVIDIA GeForce RTX 3090.}
    \label{Overall Performance}
\end{figure}

One key challenge is to deal with ambiguous semantic information during inference, which can hinder reliable decision-making. For instance, critical environmental features may be partially captured due to sensor noise or occlusion, leading to gaps in contextual understanding. Therefore, recalibrating feature channels could help ADS reconstruct missing context. Another important challenge lies in the varying relevance of driving tasks across different scenarios. For instance, mapping task aids curved-road planning but is less useful for straight paths. This suggests that fully activating all tasks is unnecessary, as different scenarios demand different tasks. Therefore, dynamic task selection might minimize the interference between driving tasks, achieving optimal planning and reduce inference~latency. 

To selectively amplify task-critical features and dynamically choose relevant tasks, we introduce \tool, a novel framework that integrates the lightweight efficiency of the Mixture of Experts (MoE) paradigm~\cite{shazeer01} into ADSs. While Mixture-of-Experts (MoEs) have shown promise across domains, their adoption in autonomous driving remains limited. Particularly, minor input variations in dynamic scenes can destabilize expert activation, and most prior works apply MoEs solely within the planning module, often using shared low-rank experts without task-specific specialization. For instance, ARTEMIS~\cite{feng2025artemis} incorporates MoE into an autoregressive planner to improve temporal modeling, but this comes at the cost of significant inference latency due to sequential token-wise processing. 

In contrast, \tool achieves both effectiveness and efficiency through an end-to-end MoE architecture that spans perception and planning. A Perception Adapter (PA) dynamically reweights BEV features to emphasize task-relevant semantics (e.g., pedestrians, road geometry), while a Mixture of Sparse Experts (MoSE) enables efficient and stable expert activation based on the driving context. MoSE customizes expert realization using sparse attention over long-term historical features, reducing computation while preserving behavioral diversity. Our framework supports task adaptivity, enhances generalization to unseen scenarios, and enables low-latency decision making, which offers a superior trade-off compared to planning-only MoE frameworks.

We integrate the \tool framework with three state-of-the-art vision-only ADSs (\ie, UniAD~\cite{hu2023planning}, VAD~\cite{jiang2023vad} and VADv2~\cite{chen2024vadv2}) and conduct experiments to evaluate the effectiveness of \tool. As depicted in Fig.~\ref{Overall Performance}, ADSs~with our \tool framework achieve reductions in collision rates by about 20\%  while reducing inference latency by approximately 25\%. Moreover, \tool shows improved multi-skill planning capabilities in certain scenarios. We also conduct a new-city generalization study to validate the generalization capability of \tool and a case study to visualize the expert pathways engaged in decision making. 

This work makes the following main contributions.
\begin{itemize}
    \item We propose a Perception Adapter (PA) to amplify task-critical features and a Mixture of Sparse Experts (MoSE) to reduce the driving task interference.
    \item We propose \tool, which is a novel framework that integrates the lightweight efficiency of the Mixture of Experts (MoE) framework into ADSs.
    \item We integrate \tool into three state-of-the-art vision-only ADSs, and conduct large scale experiments to show improved planning effectiveness and lower inference latency across both open-loop and closed-loop datasets. 
\end{itemize}

\section{Preliminary and Related Work}
\label{sec:related_work}

We first introduce the preliminary on modular end-to-end ADSs, and then discuss and review the relevant work.

\subsection{Preliminary}\label{Preliminary}
In a typical modular ene-to-end ADS pipeline, sequential modules work together for perception, prediction and planning. First, multi-view images are processed by a BEV encoder~\cite{prakash2021multi, zhang2022beverse, li2022bevformer, liang2022bevfusion, liu2023bevfusion} to generate BEV features, which are essential for understanding the environment. The perception~module then utilizes these BEV features for tasks like multi-object tracking~\cite{zeng2022motr, zhang2022mutr3d, yang2022quality, li2022time3d, hu2022monocular} and online mapping~\cite{kirillov2019panoptic, kim2020video, li2022panoptic}, providing a precise understanding of the surroundings. Next, the prediction module performs tasks like ego-state estimation, environmental interaction modeling, and navigation execution to forecast the future trajectories of ego vehicle and surrounding objects (\ie, vehicles and pedestrians). Finally, the planning module \cite{casas2021mp3, ye2023fusionad, wang2021learning, zhan2017spatially} synthesizes perceptual and predictive inputs to generate a collision-free trajectory.

\begin{figure*}[!t]
    \centering
    \includegraphics[width=0.9\textwidth]{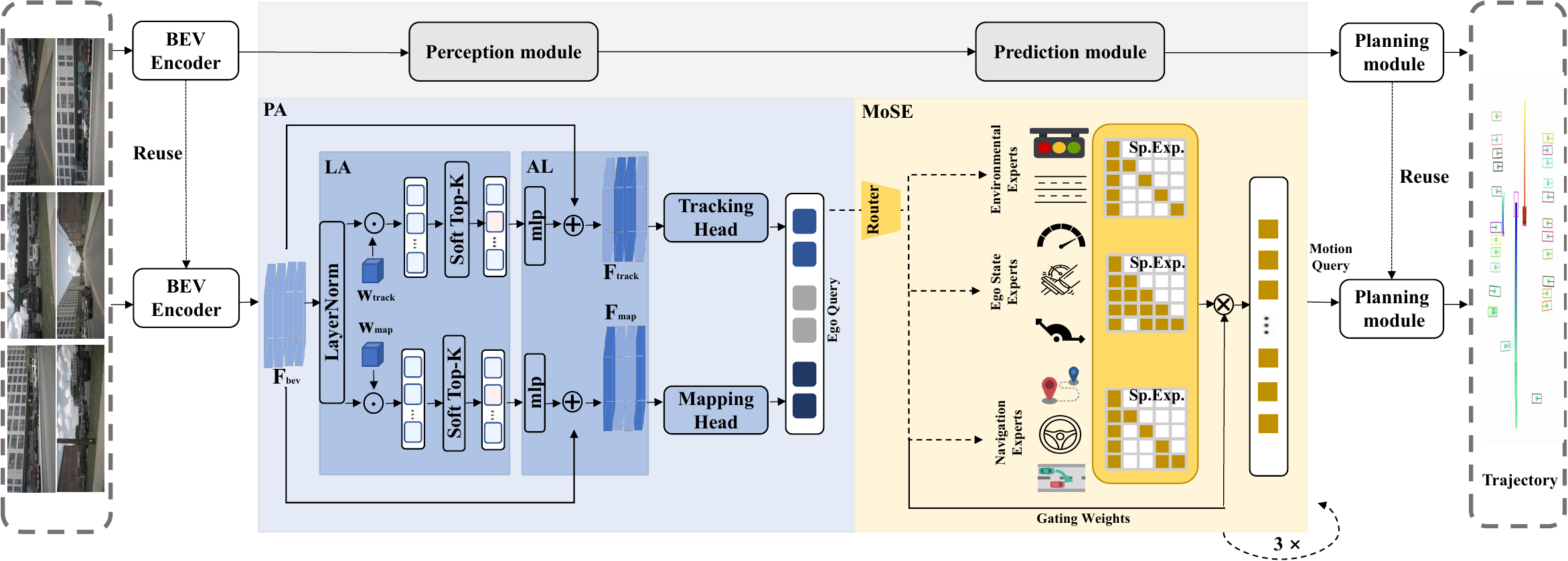}
    \caption{\textbf{Overall architecture of \tool.} \tool is built upon ADS models by retaining all the original modules except for the perception and prediction modules. The perception module is restructured as the Perception Adapter (PA) to amplify task-critical features, enhancing scene understanding. The prediction module is transformed into the Mixture of Sparse Experts (MoSE), minimizing interference among driving tasks and improving overall open-loop planning performance.}
    \label{img:overview}
    
\end{figure*}

\subsection{Related Work}
\textbf{End-to-End Autonomous Driving.} 
Early methods~\cite{pomerleau1988alvinn} use neural networks to process video and laser range data, generating steering commands via imitation learning~\cite{xiao2023scaling,zhang2021end, chen2020learning} and reinforcement learning~\cite{toromanoff2020end,chen2021interpretable}. Recent works~\cite{shao2023reasonnet, sun2023interpretable, liu2023vectormapnet, li2025hydra, jia2025drivetransformer} enhance modular end-to-end frameworks by incorporating intermediate tasks and advanced architectures for joint optimization. ST-P3~\cite{hu2022st} integrates perception, prediction and planning modules for richer features, while UniAD~\cite{hu2023planning} unifies these modules through a planning-oriented pipeline to eliminate accumulative errors. VAD~\cite{jiang2023vad} employs vectorized representations for efficient planning, and VADv2~\cite{chen2024vadv2} introduces probabilistic modeling for safer planning. Recently, large language models and world models have attracted attention for their improved planning effectiveness and strong generalization capabilities. LLMs/MLLMs in autonomous driving~\cite{jiang2024senna, shao2024lmdrive,xu2024drivegpt4,sima2023drivelm,renz2024carllava} provide textual descriptions as well as control signals for diverse driving scenarios. World models~\cite{wang2025diffad,wang2024driving,wang2023drivedreamer,yang2023bevcontrol,zhang2024bevworld,gao2023magicdrive} build comprehensive environment representations and make planning decisions based on predicted future states. Despite their promise, high inference latency remains a key barrier to real-world deployment. Even when LLMs or world models are deployed, vehicles often retain a conventional end-to-end model~\cite{tian2024drivevlm,chen2025solve} as a fallback to ensure robust execution. To reduce inference time of end-to-end ADSs, DriveAdapter~\cite{jia2023driveadapter} adopts a teacher-student paradigm where the student model gains driving knowledge from the teacher model and aligns features between privileged and raw sensor inputs. PlanKD~\cite{feng2024road} further enables student planner to inherit planning-relevant knowledge to improve performance. However, these two approaches purse the lower inference latency at the cost of planning effectiveness. 
Our work aims to improve both planning effectiveness and inference efficiency of modular end-to-end ADSs with a MoE architecture.

\textbf{Mixture-of-Experts (MoE).}
MoE models aim to scale model capacity while reducing computational costs by activating only a subset of parameters for each input. Recent works utilize MoE in large language models~\cite{du2022glam,team2023llama,jiang2024mixtral} and vision models~\cite{daxberger2023mobile,riquelme2021scaling} to improve training efficiency and inference performance. In autonomous driving, some approaches~\cite{ohn2020learning,nazeri2021exploring} rely on supervised learning from historical driving data, with MoE facilitating the execution of behavior instructions in predefined driving scenarios. Despite their potential, these models still have unsatisfactory planning effectiveness, facing challenges in generalization abilities to adjust their behavior across unseen driving scenarios. Some MoE-based approaches have been applied to the perception module, such as camera selection~\cite{fang2020multi,morra2023mixo} and visual token selection~\cite{zhang2024minidrive}. Others focus on the planning module, targeting tasks like trajectory selection~\cite{feng2025artemis, pini2023safe} and driving intention prediction~\cite{yuan2023temporal}. While effective for specific subtasks, these methods often lack unified semantic and behavioral modeling, making them highly sensitive to noise in both perception and planning. Moreover, they typically introduce additional computational overhead during inference.
In contrast, our proposed framework, \tool, amplifies task-critical features and reduces task interference, optimizing both planning effectiveness and inference efficiency.

\section{Methodology}
\label{sec:methodology}

We design and implement \tool, a novel Mixture of Experts (MoE) framework, to improve the planning effectiveness and inference efficiency of ADSs. 

\subsection{Approach Overview}
The framework overview of \tool is shown in Fig.~\ref{img:overview}. \tool is built upon an ADS, retaining all original modules except for the perception and prediction modules. We restructure the perception module as the \textbf{Perception Adapter (PA)} to amplify task-related features for context-aware scene understanding. We transform the prediction module into the \textbf{Mixture of Sparse Experts (MoSE)} to dynamically activate driving tasks, minimizing interference. We define a combined \textbf{Training Loss} to ensure task effectiveness while promoting efficient expert utilization.

As multi-view camera images are encoded into BEV features through the BEV Encoder, the PA selects task-relevant feature dimensions via a \emph{Learned Adapter (LA)}, and amplifies task-related semantics for the tracking and mapping transformers using \emph{Alignment Layers (AL)}. The PA outputs an ego query, combining the agent query (\ie, the output of tracking transformer), the map query (\ie, the output of mapping transformer), and a learnable embedding. Prediction tasks are categorized into three groups with eight \emph{Sparse Experts (Sp.Exp.)}, each using a sparse attention mechanism based on their respective focus. The MoSE uses the ego query from the PA to dynamically select and activate the most relevant experts through a \emph{Router} based on the current driving scenario. The motion query output from the MoSE is then passed to the planning module to determine the final driving route.

\subsection{Perception Adapter}\label{BEV Features Adapter Module}
BEV features from the BEV Encoder capture various semantics (\eg, roads, vehicles, and traffic signs). However, the focus of each perception task (\ie, multi-object tracking~\cite{zeng2022motr, zhang2022mutr3d, yang2022quality, li2022time3d, hu2022monocular} and online mapping~\cite{kirillov2019panoptic, kim2020video, li2022panoptic}) differs. Directly passing all features to the tracking and mapping transformers might cause non-critical dimensions to overshadow key features, leading to information loss.~To~address~this,~we propose PA, composed of a \emph{Learned Adapter} and an \emph{Alignment Layer}, to amplify task-critical features. 

\textbf{Learned Adapter.}
It is designed to select a subset~of task-specific BEV features for each task in the perception module. Instead of relying on static feature selection methods that remove or merge features based on predefined criteria \cite{bolya2022token, rao2021dynamicvit, yin2022vit}, we adapt the idea of conditional adapter~\cite{lei2023conditional}, originally designed for token selection in fine-tuning.

Specifically, We first normalize and pool the BEV features along the temporal dimensions to obtain a frame-agnostic BEV representation $\tilde{\mathrm{BEV}}  \in \mathbb{R}^{d \times H \times W}$, where $d$ is the feature dimension and $H \times W$ is the spatial resolution from the BEV Encoder. For each task $t \in T$ (\eg, tracking, mapping), we introduce a task-specific learnable parameter $w^{(t)}\in \mathbb{R}^{d}$ to dynamically weight feature dimensions. The initial task-aware selection score $s^{(t)} \in \mathbb{R}^{d}$ is computed by Eq.~\ref{q1}, where $\tilde{\mathrm{BEV}}_{:,i,j}$ denotes the $d$-dimensional feature vector at spatial position $(i,j)$. This score reflects the importance of each feature dimension for the given task.
\begin{equation}
    \begin{aligned}
        s = \frac{1}{H \times W}\sum_{i,j} \tilde{\mathrm{BEV}}_{:,i,j} \odot w
    \end{aligned}
    \label{q1}
\end{equation}

Based on the task-specifc score $s^{(t)}$, we compute the soft channel selection weights $\lambda^{(t)} \in [0, 1]^{d}$ by solving the constrained optimization problem defined in Eq.~\ref{lambda_},
\begin{equation}
    \begin{aligned}
        f(s) := &\ \arg\max_{\lambda} \, (s)^\top \lambda + \epsilon \Omega(\lambda) \\
        {s.t.}\ &\ \mathbf{1}^\top \lambda =  \tau, \lambda \in [0, 1]^{d}
    \end{aligned}
    \label{lambda_}
\end{equation}
where the entropy $\Omega(\lambda)$ regularizes the solution of $\lambda$, promoting smoothness in the feature selection process. The constant $\tau$ represents the number of key feature dimensions. The constraint $\mathbf{1}^\top \lambda =  \tau$ ensures the model focuses on $\tau$ dominant channels and the open-interval constraint $\lambda \in [0, 1]^{d}$ enables continuous weighting of channels. Since this optimization problem does not have a closed-form solution for $\epsilon > 0$ and $\tau > 1$, we employ the iterative algorithm by Lei \etal~\cite{lei2023conditional} to get subdifferentiable soft weights for $\lambda$. 

\textbf{Alignment Layer.} It is designed to reweight feature dimensions to recalibrate and amplify task-critical features.

Specifically, given the selection weight $\lambda^{(t)}$ from the \emph{Learned Adapter}, we obtain the calibrated features $F_{align}$ by Eq.~\ref{bev:output}, 
\begin{equation}
    F_{align} = MLP(BEV \odot \lambda) + BEV
    \label{bev:output}
\end{equation}
This equation combines BEV features from the BEV Encoder with amplified features. The term $BEV \odot \lambda$ represents element-wise multiplication, effectively amplifying certain features based on the learned weight distribution. The $MLP$ layer further introduces non-linearity to the transformed features. The result is added back to the original BEV features to generate $F_{align}$, which not only preserves the original spatial information but also creates a shortcut path for gradients during backpropagation.

The aligned features $F_{align}$ are then processed by the tracking transformer and mapping transformer to generate agent and map queries, providing a comprehensive understanding of the surroundings. Finally,~the~two queries are concatenated with a learnable embedding to form an ego query $\mathcal{F}_{ego} \in \mathbb{R}^{B \times L \times d}$, where $B$ is the batch size, $L$ is the sequence length and $d$ is the feature dimension. This ego query serves as the input to the MoSE, allowing the model to process the concatenated information for further tasks.

\subsection{Mixture of Sparse Experts }\label{Expert Router Module}
The prediction module includes various driving tasks to enhance ADS decision-making. However, interactions among these tasks, especially with long historical sequences, can cause interference and raise computational cost. To address this, we propose MoSE, which uses \emph{Sparse Experts} and a \emph{Router} to dynamically activate tasks based on the ego query $\mathcal{F}_{ego}$ from the previous module.

\textbf{Sparse Experts.}
To forecast future trajectories of the ego vehicle and nearby agents, the prediction module coordinates three critical tasks: ego-state estimation, environmental interaction modeling and navigation execution. These are formalized into three expert categories, \ie, \emph{Environmental Experts}, \emph{Ego State Experts} and \emph{Navigation Experts}, with a total of eight experts.

\emph{Environmental Experts} process environmental information provided by the perception module, such as traffic signals, lane markings, and obstacles. Specifically, the \emph{Tracking Expert} focuses on modeling dynamic foreground objects, while the \emph{Mapping Expert} maintains map topology with lane-graph connectivity. 

\emph{Ego State Experts} regulate vehicle dynamics based on historical trajectories. Specifically, the \emph{Velocity Expert},~\emph{Yaw Expert} and \emph{Acceleration Expert} collaborate to process vehicle-specific information for smooth trajectories.

\emph{Navigation Experts} execute hierarchical route planning for vehicles. Specifically, The \emph{Reference Point Expert} aligns trajectories with reference points, the \emph{BEV Expert} provides spatial context, and the \emph{Command Expert} interprets and executes navigation commands to adjust vehicle behavior.

Each expert is equipped with a sparse attention mechanism $MHCA()$ defined by Eq.~\ref{sparse attention} based on respective focus, 
\begin{equation}
    MHCA(Q, K, V) = \sum_{j \in C_{i}}softmax\left(\frac{Q_i K_j^T}{\sqrt{d_k}}\right) V_j
    \label{sparse attention}
\end{equation}
where $Q_i$ denotes the $i$-th query vector, $K_j$ and $V_j$ represent the selected keys and values, and $C_i$ defines the sparse computation for the $i$-th query. 

Specifically, \emph{Environmental Experts} adopt block-wise sparse attention, \ie, $C_i = \{ j \mid \left\lfloor j/m \right\rfloor = \left\lfloor i/m \right\rfloor \}$, restricting attention to local regions with a block size of $m$, allowing efficient spatial pattern capture within environmental regions (\eg, obstacles and traffic signs) while reducing computational costs. \emph{Ego State Experts} utilize a sliding window attention mechanism with a window size of $w$, where $C_i = [i - w, i + w]$, enabling rapid adaptation to changes in ego state (\eg, speed, and acceleration) while maintaining manageable computational complexity. \emph{Navigation Experts} adopt a global TopK attention, defined by  $C_i = \{ j \mid K_j \in \text{Top}_k(Q_i K^T) \}$, selecting key positions with the highest attention scores for the current query to capture long-range dependencies and prioritize critical information across the entire scene, such as waypoints and navigation cues.

Each expert is tailored to the specific task required by the prediction module and encodes domain-specific data through a fully connected layer with modality-specific normalization. Specifically, navigation experts embed routing instructions as text features, environmental experts extract geometric features from perception outputs, and ego-state experts represent vehicle dynamics as time-series data. These embeddings, denoted as $\mathcal{F}_{expert}$, are fused with the ego query $\mathcal{F}_{ego}$ from the Perception Adapter (PA) via the sparse attention mechanism $MHCA()$ (Eq.~\ref{interaction}), yielding expert-informed queries $\bar{\mathcal{F}}_{expert}$. This modular design facilitates efficient integration of diverse inputs into a unified predictive framework.
\begin{equation}
    \bar{\mathcal{F}}_{expert} = MHCA \left( \mathcal{F}_{ego}, \mathcal{F}_{expert}, \mathcal{F}_{expert}\right)
    \label{interaction}
\end{equation}

\textbf{Router.} Each expert in the prediction module generates a $\bar{\mathcal{F}}_{expert}$. We hope to activate different experts based on the various driving scenarios. The \emph{Router} is responsible for weighting the top-$k$ experts through logits to dynamically select experts rather than fully activating all experts.

To compute expert logits, we adopt the gating mechanism from MoE~\cite{shazeer01} and introduce a learnable parameter $ \mathbf{W}_{\text{gate}} \in {R}^{d \times N}$ , where $d$ is the feature dimension and $N$ denotes the number of experts, mapping $\mathcal{F}_{ego}$ to relevant experts over a batch. To introduce stochasticity, Gaussian noise $\eta \sim \mathcal{N}(0,1)$ is added to the logits during training, scaled by a learnable matrix $\mathbf{W}_{\text{noise}}$ of the same shape as $\mathbf{W}_{\text{gate}}$. The logits are then computed using $\mathcal{F}_{ego}$ by Eq.~\ref{gating},
\begin{equation}
    \begin{aligned}
        g(x) = x \cdot \mathbf{W}_{\text{gate}} &+ \eta \cdot \left( \text{softplus}(x \cdot \mathbf{W}_{\text{noise}}) + \epsilon \right) \\
        &\quad \eta \sim \mathcal{N}(0,1), \quad \epsilon > 0
    \end{aligned}
    \label{gating}
\end{equation}
where $\eta \sim \mathcal{N}(0,1)$ and $\epsilon > 0$ is a positive factor controlling the noise scale. This noise-enhanced gating mechanism enables dynamic expert selection based on input features while incorporating randomness to prevent overfitting.

Then, we introduce a function $\mathcal{R}(\cdot)$, defined in Eq.~\ref{Router}, to select the top-$k$ experts based on the loigts of experts.
\begin{equation}
    \mathcal{R}(x) = {TopK}({softmax}(g(x)), k)
    \label{Router}
\end{equation}
Taking $\mathcal{F}_{ego}$ as input, the result of $\mathcal{R}(\mathcal{F}_{ego})$ represents the routing scores of the top-$k$ experts. The gating function $\mathcal{R}(\cdot)$ employs a decoupled selection mechanism based on high-level scene understanding, avoiding information from experts that could introduce noise.

Finally, we compute the motion query $\mathcal{F}_{Motion}$ of the selected top-$k$ experts by Eq.~\ref{MoE},
\begin{equation}
    \mathcal{F}_{Motion} = \sum_{i=1}^{k} {\mathcal{R}(\mathcal{F}_{ego})}_{i} \cdot \bar{\mathcal{F}}_{expert_i} \\
    \label{MoE}
\end{equation}
where $\mathcal{R}(\mathcal{F}_{ego})_{i}$ and $\bar{\mathcal{F}}_{expert_i}$ respectively represent the routing score and the expert-specific query for the selected expert~$i$. The resulting $\mathcal{F}_{Motion}$ integrates and reweights ego features, environmental features and navigation features, providing a global feature representation tailored to the current driving scenario for planning. 

\subsection{Training Loss}\label{training loss}
Shazeer \etal~\cite{shazeer01} has observed that models tend to focus higher weights on a few experts, leading to underutilization of others and inefficient task distribution. To address this, we incorporate a switch-loss \cite{fedus2022switch} to encourage more balanced weight allocation across the experts. 

Switch-loss balances the allocation of queries over a batch across $N$ experts by penalizing experts that receive disproportionately high routing probabilities by Eq.~\ref{switch loss},
\begin{equation}
    \mathcal{L}_{switch} = N \cdot \sum_{i=1}^{N}  f_i  \cdot  \mathcal{P}_{i} 
    \label{switch loss}
\end{equation}
where $f_i$ is the actual query load on expert $i$, while $\mathcal{P}_{i}$ is its expected routing probability. Switch-Loss penalizes experts that are overused relative to their expected share.

The final model loss, defined by Eq.~\ref{total loss}, incorporates the losses from all tasks along with the switch loss,
\begin{equation}
    \label{total loss}
    \begin{aligned}
        \mathcal{L}_{\text{total}} = & \, \alpha_1 \mathcal{L}_{\text{perception}} + \alpha_2 \mathcal{L}_{\text{prediction}} \\
        & + \alpha_3 \mathcal{L}_{\text{planning}} + \alpha_4 \mathcal{L}_{\text{switch}},
    \end{aligned}
\end{equation}
where $\mathcal{L}_{\text{perception}}$, $\mathcal{L}_{\text{prediction}}$ and $\mathcal{L}_{\text{planning}}$ respectively represent the loss functions for the perception, prediction and planning modules in the baseline ADSs.
\section{Experiments}
\label{sec:experiments}

We first introduce the implementation of our experiments. Then, we report the overall performance and modular performance of \tool. Finally, we conduct an ablation study, a generalization evaluation and a qualitative study. 

\subsection{Implementation Details}\label{subsec:experiments_setup}

\textbf{Dataset.} We evaluate \tool using both open-loop and closed-loop datasets.
For open-loop evaluation, we adopt the nuScenes dataset~\cite{Caesar2019nuScenesAM}, which contains 1,000 real-world scenes with sensor data from 6 cameras, 1 LiDAR, 5 radars, GPS, and IMU. We use only the 6 camera images as visual input.
For closed-loop evaluation, we use the Bench2Drive benchmark~\cite{jia2024bench} with 2 million annotated training frames from 12 towns and 23 weather conditions in CARLA~\cite{Dosovitskiy17}. Its evaluation set includes 220 routes across 44 interactive scenarios (denoted as Bench2Drive220).

\textbf{Baselines.} Our work is orthogonal to existing transformer based frameworks and can be integrated into any of them. To evaluate the effectiveness and efficiency of \tool, we adopt three state-of-the-art models, \ie, UniAD~\cite{hu2023planning}, VAD~\cite{jiang2023vad}, and VADv2~\cite{chen2024vadv2}, as representative and publicly available baselines.

\textbf{Training.} We instantiate \tool into UniAD, VAD and VADv2, denoted as Expert-UniAD, Expert-VAD, and Expert-VADv2. All approaches are trained using the same hyperparameters as respective baselines for fair comparison. Experiments are run on 8 NVIDIA Tesla A100 GPUs.

\textbf{Metrics.} Following previous works, we use ST-P3~\cite{hu2022st} metrics, \ie, L2 errors and collision rates, to evaluate open-loop driving effectiveness of ADSs. For closed-loop evaluation, we adopt Success Rate (SR), Route Completion (RC), and Driving Score (DS)~\cite{Dosovitskiy17} to assess effectiveness. We further report Latency (average inference time per forward pass), GFLOPs (computational complexity), and Params (model size) as efficiency metrics. Paired t-tests across benchmarks and metrics demonstrate that ExpertAD achieves statistically significant improvements over all baseline ADSs, with an average p-value of 0.026 ($p < 0.05$). All results are averaged over five independent runs.

\subsection{Joint Results}\label{subsec:joint_results}

\begin{table}[!t]
    \centering
    \small
    \renewcommand{\arraystretch}{1.3}
    \begin{adjustbox}{width=\linewidth}
    \begin{tabular}{l c c c c c c c c}
    \toprule
    \multirow{2}{*}{\textbf{Approach}} 
    & \multicolumn{2}{c}{\textbf{Open-loop Metric}} 
    & \multicolumn{3}{c}{\textbf{Closed-loop Metric}} 
    & \multicolumn{3}{c}{\textbf{Efficiency Metric}} \\
    
    \cmidrule(lr){2-3} \cmidrule(lr){4-6} \cmidrule(lr){7-9}
    
    & Avg.col ↓ & Avg.L2 ↓ & DS ↑ & SR ↑ & RC ↑ & Latency ↓ & GFLOPs ↓ & Params ↓ \\
    
    \midrule
    UniAD                    & 0.31 &   1.03     & 44.62 & 14.09 & 68.68 & $534 \pm 18$ ms & $\sim 856$ & $\mathbf{\sim 89M}$ \\
    \rowcolor[gray]{0.9}\textbf{Expert-UniAD}  & \textbf{0.24} &   \textbf{0.89}     & \textbf{55.49} & \textbf{20.63} & \textbf{81.04} & $\mathbf{445 \pm 20}$ ms & $\mathbf{\sim 728}$ & $\sim 125M$ \\
    VAD                      & 0.43 &    \textbf{1.21}    & 43.31 & 17.27 & 61.60 & $225 \pm 25$ ms & $\sim 558$ & $\mathbf{\sim 58M}$ \\
    \rowcolor[gray]{0.9}\textbf{Expert-VAD}    & \textbf{0.34} &   \textbf{1.10}     & \textbf{52.53} & \textbf{19.53} & \textbf{76.73} & $\mathbf{157 \pm 23}$ ms & $\mathbf{\sim 461}$ & $\sim 90M$ \\
    VADv2                    & 0.12 &   0.33     & 75.90 & 55.01 & \textbf{90.08} & $330 \pm 18$ ms & $\sim 660$ & $\mathbf{\sim 76M}$ \\
    \rowcolor[gray]{0.9}\textbf{Expert-VADv2}  & \textbf{0.10} &    \textbf{0.28}    & \textbf{78.18} & \textbf{58.34} & 89.32 & $\mathbf{258 \pm 22}$ ms & $\mathbf{\sim 573}$ & $\sim 105M$ \\
    
    \bottomrule
    \end{tabular}
    \end{adjustbox}
    \renewcommand{\arraystretch}{1.0}
    \caption{\textbf{Overall performance comparison.} \tool achieves improved planning effectiveness and lower inference latency compared to baseline models. Performance is measured on a single NVIDIA GeForce RTX 3090 GPU.}
    \label{joint results}
\end{table}

\begin{table}[!t]
    \centering
    \small
    \renewcommand{\arraystretch}{1.3} 
    \begin{adjustbox}{width=\linewidth}
    \begin{tabular}{lcccccccccc}
    \hline
    Approach              & \multicolumn{2}{c}{Merge ↑} & \multicolumn{2}{c}{Overtake ↑} & \multicolumn{2}{c}{EmgBrake ↑} & \multicolumn{2}{c}{GiveWay ↑} & \multicolumn{2}{c}{Tsign ↑} \\ \hline
    UniAD                 & \multicolumn{2}{c}{12.66}      & \multicolumn{2}{c}{13.33}         & \multicolumn{2}{c}{20.00}         & \multicolumn{2}{c}{10.00}        & \multicolumn{2}{c}{13.23}      \\
    \rowcolor[gray]{0.9}\textbf{Expert-UniAD} & \multicolumn{2}{c}{\textbf{27.38}}      & \multicolumn{2}{c}{\textbf{23.67}}         & \multicolumn{2}{c}{\textbf{51.67}}         & \multicolumn{2}{c}{\textbf{20.00}}        & \multicolumn{2}{c}{\textbf{40.93}}      \\ \hline
    VAD                   & \multicolumn{2}{c}{8.89}      & \multicolumn{2}{c}{20.44}         & \multicolumn{2}{c}{18.64}         & \multicolumn{2}{c}{20.00}        & \multicolumn{2}{c}{18.66}      \\
    \rowcolor[gray]{0.9}\textbf{Expert-VAD}   & \multicolumn{2}{c}{\textbf{22.25}}      & \multicolumn{2}{c}{\textbf{26.38}}         & \multicolumn{2}{c}{\textbf{49.33}}         & \multicolumn{2}{c}{\textbf{20.00}}        & \multicolumn{2}{c}{\textbf{51.53}}      \\ \hline
    VADv2                 & \multicolumn{2}{c}{36.25}      & \multicolumn{2}{c}{48.33}         & \multicolumn{2}{c}{74.28}         & \multicolumn{2}{c}{\textbf{50.00}}        & \multicolumn{2}{c}{60.14}      \\
    \rowcolor[gray]{0.9}\textbf{Expert-VADv2} & \multicolumn{2}{c}{\textbf{40.44}}      & \multicolumn{2}{c}{\textbf{48.33}}         & \multicolumn{2}{c}{\textbf{78.42}}         & \multicolumn{2}{c}{40.00}        & \multicolumn{2}{c}{\textbf{65.78}}      \\ \hline
    \end{tabular}
    \end{adjustbox}
    \renewcommand{\arraystretch}{1.0} 
    \caption{\textbf{Multi-skill capabilities in rare scenarios.} ExpertAD shows improved overall performance in Overtake, Merge, and T-sign scenarios, while maintaining comparable performance in EmgBrake and Giveway scenarios.}
    \label{Multi-skill capabilities}
\end{table}

As shown in Table~\ref{joint results}, \tool consistently enhances both planning effectiveness and inference efficiency across all baselines. Compared to UniAD, our method achieves a 23\% reduction in collision rates, a 14\% decrease in L2 errors, and a 1.2× speedup. Expert-VAD and Expert-VADv2 exhibit similar improvements, reducing collision rates by 21\% and 17\%, lowering L2 errors by 9\% and 15\%, and achieving 1.4× and 1.3× speedups, respectively. In closed-loop evaluation, ExpertAD further improves DS, SR, and RC by 16\%, 22\%, and 14\%, averaged over the three ADSs. \tool incurs minimal parameter overhead while significantly lowering FLOPs, achieving a balanced trade-off between accuracy and efficiency. Unlike prior methods that favor speed at the cost of planning quality, \tool offers joint gains in both. 

Bench2Drive evaluates ADSs in five scenario groups. As shown in Table~\ref{Multi-skill capabilities}, ExpertAD excels in Emergency Braking situations (\eg, PedestrianCrossing, ParkingCutIn), Merging situations (\eg, HighwayCutIns, JunctionTurns), and Traffic Signs (\eg, StopSigns, TrafficLights), where rich perceptual information from pedestrians, crossroads, and traffic lights provides critical context. However, for Overtaking scenarios (\eg, Accident, ConstructionZones) and Giving Way scenarios (\eg, YieldingtoEmergencyVehicles), ExpertAD shows only marginal or inconsistent improvements. These scenarios demand complex, human-like reasoning, suggesting that rule-based fallback systems are still needed.

\subsection{Modular Results}\label{subsec:modular_results}

\begin{table}[!t]
    \centering
    \scriptsize
    \begin{adjustbox}{width=0.8\linewidth}
    \begin{tabular}{lcccccc}
    \toprule
    Approach              & $\tau$ & DS ↑  & SR ↑  & RC ↑  \\
    \midrule
    UniAD        & -     &  44.62   &  14.09   &  68.68   \\
    UniAD + PA & 32 &   40.35   &  15.00  &   58.94   \\
    UniAD + PA & 64 &  48.25    &  16.93  &   70.47   \\
    \rowcolor[gray]{0.9} \textbf{UniAD + PA} & \textbf{128} &  \textbf{52.53}   &  \textbf{18.41}  &  \textbf{76.73}   \\
    UniAD + PA & 256 &   43.31   &  16.50  &  64.83    \\
    \bottomrule
    \end{tabular}
    \end{adjustbox}
    \caption{\textbf{PA results.} UniAD with PA achieves its best results with $\tau = 128$.}
    \label{Perception}
\end{table}

\begin{table}[!t]
    \centering
    \small
    \begin{adjustbox}{width=0.8\linewidth}
    \begin{tabular}{lcccccc}
    \toprule
    Approach              & Top-K & DS ↑  & SR ↑  & RC ↑  \\
    \midrule
    UniAD        & -     &  44.62   &  14.09   &  68.68   \\
    UniAD + MoSE & Top-8 &  46.24    &  17.27  &  67.54    \\
    \rowcolor[gray]{0.9}\textbf{UniAD + MoSE} & \textbf{Top-4} &  \textbf{49.32}   &  \textbf{18.41}   &  \textbf{72.03}   \\
    \bottomrule
    \end{tabular}
    \end{adjustbox}
    \caption{\textbf{MoSE results.} Top-4 experts contribute to the best planning performance over baseline UniAD.}
    \label{Top-N}
\end{table}

Following the sequential order of the PA and MoSE module in~\tool, we evaluate the effectiveness of each module in comparison to the baseline models.

\begin{table}[!t]
    \centering
    \scriptsize
    \renewcommand{\arraystretch}{0.75}
    \begin{adjustbox}{width=\linewidth}
    \begin{tabular}{lccccc}
    \toprule
    Approach & MLP() & ADD() & AMOTA↑ & AMOTP↓ \\ 
    \midrule
    \textbf{UniAD} & & & 0.388 & 1.304 \\ 
    \midrule
    \multirow{3}{*}{\textbf{Expert-UniAD}} & \ding{51} & & 0.384 & 1.306 \\ 
    & & \ding{51} & {0.390} & {1.298} \\ 
    & \ding{51} & \ding{51} & \textbf{0.404} & \textbf{1.277} \\ 
    \bottomrule
    \end{tabular}
    \end{adjustbox}
    \renewcommand{\arraystretch}{1.0}
    \caption{\textbf{Ablation results for PA components.} PA with an MLP layer introduces non-linearity to transformed features, while PA using addition function preserves feature stability.}
    \label{tab: Ablation Study for PA components.}
\end{table}

\begin{table*}[!t]
    \centering
    \scriptsize
    \renewcommand{\arraystretch}{0.8}
    \begin{tabular}{lllccc>{\columncolor[gray]{0.9}}cccc>{\columncolor[gray]
    {0.9}}cl}
    \toprule
    \multirow{2}{*}{{Approach}} & \multirow{2}{*}{Router} & \multirow{2}{*}
    {Sparse Attention}& \multicolumn{4}{c}{L2 (m) ↓} & \multicolumn{4}{c}{Col. Rate (\%) ↓} &
    \multirow{2}{*}{Latency ↓} \\
    \cmidrule(lr){4-7} \cmidrule(lr){8-11}
    &&& 1 s & 2 s & 3 s & Avg. & 1 s & 2 s & 3 s & Avg. \\
    \midrule
    \multirow{4}{*}{\textbf{Expert-UniAD}} &&& 0.46 & 0.91 & 1.57 & 0.98 & 0.05 &
    0.20 & 0.55 & 0.27 & $623 \pm 35 ms$ \\
    &\multicolumn{1}{c}{\ding{51}}&& \textbf{0.34} & \textbf{0.76} & 1.47 & \textbf{0.86} & \textbf{0.03} & \textbf{0.16}
    & 0.45 & \textbf{0.21} & $591 \pm 30 ms$\\
    &&\multicolumn{1}{c}{\ding{51}}& 0.46 & 0.95 & 1.64 & 1.02 &0.05 & 0.29 & 0.52 &
    0.29 & $476 \pm 25 ms$\\
    &\multicolumn{1}{c}{\ding{51}}&\multicolumn{1}{c}{\ding{51}}& 0.42 &
    0.85 & \textbf{1.41} & 0.89 &0.07 & 0.22 & \textbf{0.43} &
    0.24 & $\mathbf{445 \pm 20 ms}$\\
    \midrule
    \multirow{4}{*}{\textbf{Expert-VAD}} &&& 0.58 & 1.10 & 1.69 & 1.12 & 0.07 & 0.22
    & 0.79 & 0.36 & $289 \pm 28 ms$\\
    &\multicolumn{1}{c}{\ding{51}}&& 0.53 & \textbf{1.05} & \textbf{1.67} & \textbf{1.08} & \textbf{0.06} &
    \textbf{0.16} & \textbf{0.73} & \textbf{0.32} & $256 \pm 22 ms$\\
    &&\multicolumn{1}{c}{\ding{51}}& 0.62 & 1.14 & 1.70 & 1.15 & 0.12 & 0.28 & 0.84
    & 0.41 & $180 \pm 20 ms$ \\
    &\multicolumn{1}{c}{\ding{51}}&\multicolumn{1}{c}{\ding{51}}& \textbf{0.53} &
    1.07 & 1.69 & 1.10 & 0.07 & 0.22 &
    0.74 & 0.34 & $\mathbf{157 \pm 23 ms}$ \\
    \bottomrule
    \end{tabular}
    \renewcommand{\arraystretch}{1.0}
    \caption{\textbf{Ablation results for MoSE components.} MoSE with Router routes the ego query to the top-4 experts, while MoSE without Router activates all experts. All experiments are conducted using PA. Router notably enhances planning metrics, while Sparse Attention significantly reduces latency.}
    \label{Ablation_Study}
\end{table*}

\textbf{PA Results.}
We evaluate the hyperparameter $\tau$, which controls the number of selected BEV features in the PA module. As shown in Table~\ref{Perception}, performance improves as $\tau$ increases, peaking at $\tau = 128$ with gains of 17\% (DS), 30\% (SR), and 11\% (RC) over the baseline. Larger $\tau$ values capture richer features, while overly large values (e.g., $\tau = 256$) may introduce redundancy.

\textbf{MoSE Results.}
We compare three UniAD settings, baseline (no expert activation), Top-8 (all experts), and Top-4 (sparse activation). As shown in Table~\ref{Top-N}, Top-8 achieves modest DS gains, confirming the benefit of expert specialization. However, activating all experts leads to cautious decisions that reduce violations but slightly lower route completion. Top-4 further improves DS, SR, and RC by 11\%, 30\%, and 5\%, respectively, as MoSE effectively reduces interference by selecting domain-relevant experts.

\begin{figure}[!t]
    \centering
    \begin{subfigure}{1.0\linewidth}
    \includegraphics[width=1.0\linewidth]{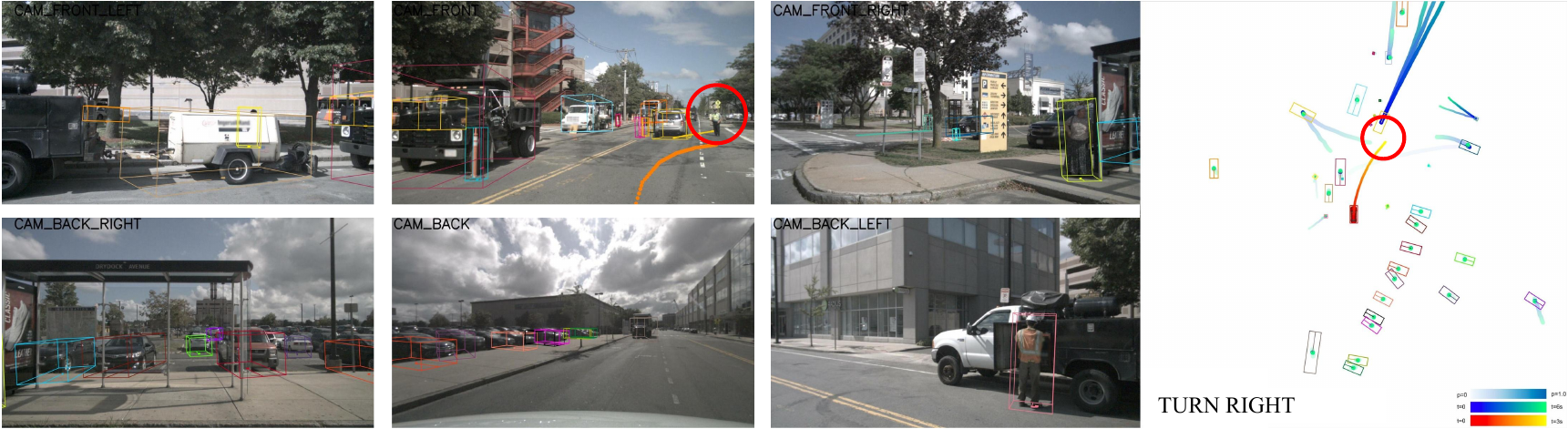}
    \caption{visualization result of UniAD}
    \label{Perception results-a}
    \end{subfigure}

    \begin{subfigure}{1.0\linewidth}
    \includegraphics[width=\linewidth]{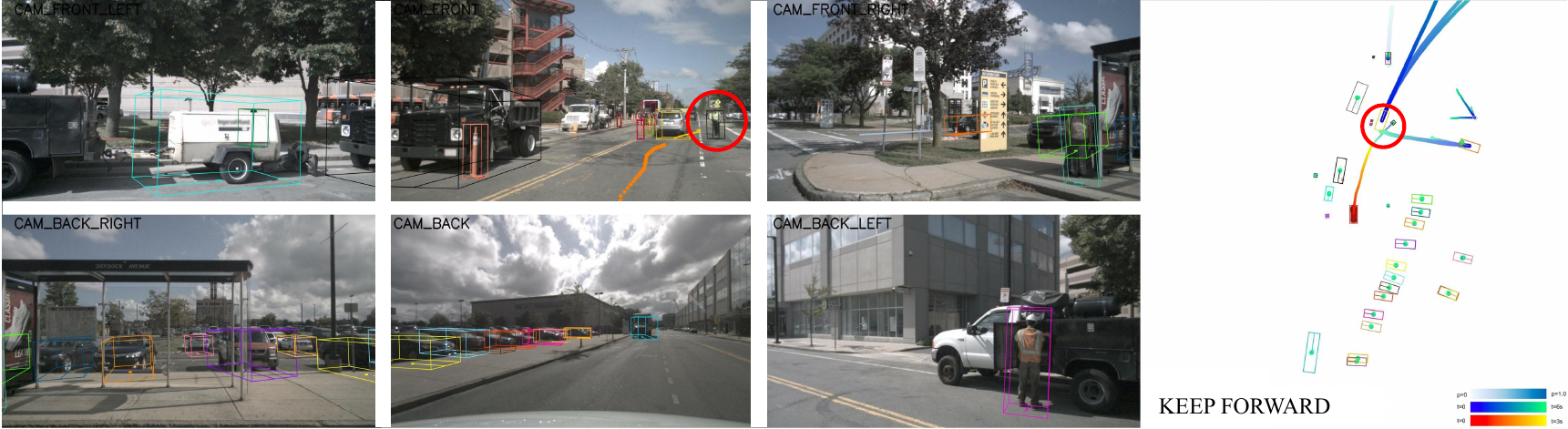}
    \caption{visualization result of Expert-UniAD}
    \label{Perception results-b}
    \end{subfigure}

    \caption{\textbf{\tool recovery visualization.} 
    (a) UniAD misses a traffic officer on the right, leading the vehicle to drift toward them. (b) Expert-UniAD detects the officer, adjusts the trajectory, and completes the lane change safely.}
    \label{Perception results}
\end{figure}

\subsection{Ablation Study}\label{subsec:ablation_study}
We conduct ablation study on the PA and MoSE module respectively in Expert-UniAD and Expert-VAD. As VADv2 extends VAD with probabilistic modeling in planning, we exclude Expert-VADv2 from this study. To avoid the runtime overhead introduced by closed-loop evaluation, following experiments are performed under the open-loop setting.

\textbf{PA Ablation.}
Since VAD does not implement a perception head, we compare only against UniAD for multi-object tracking. We adopt~standard metrics~\cite{hu2022st}, \ie, Average Multi-Object Tracking Accuracy (AMOTA) and Average Multi-Object Tracking Precision (AMOTP) to assess the effectiveness of multi-object tracking. As shown in Table~\ref{tab: Ablation Study for PA components.}, removing the addition function reduces performance below the baseline, while adding the MLP layer notably improves AMOTA, highlighting its effectiveness in enhancing task-relevant features.

\textbf{MoSE Ablation.} Table~\ref{Ablation_Study} demonstrates that the Router effectively reduces L2 error and collision rate by mitigating task interference through selective expert activation. Sparse Attention further lowers latency with little or no loss in planning performance, and improves long-term prediction (e.g., 3s) by filtering out irrelevant tokens. Notably, Expert-UniAD with both Router and Sparse Attention achieves a significant latency reduction of 178ms, while improving planning effectiveness by 0.09m (L2 error) and 0.03\% (collision rate). Expert-VAD shows similar benefits with a 132ms latency reduction and gains of 0.02m and 0.02\%, confirming a favorable trade-off between efficiency and effectiveness.

\begin{table}[!t]
    \centering
    \begin{adjustbox}{width=\linewidth}
    \begin{tabular}{lcccc}
    \toprule
    \multirow{2}{*}{Approach} & \multicolumn{2}{c}{Train on Boston} & \multicolumn{2}{c}
    {Train on Singapore}\\
    \cmidrule(lr){2-3} \cmidrule(lr){4-5}
    & \multicolumn{1}{c}{Avg. L2 ↓} & \multicolumn{1}{c}{Avg. Col. ↓} & \multicolumn{1}{c}{Avg. L2 ↓} & \multicolumn{1}{c}{Avg. Col. ↓}\\
    \midrule
    \textbf{UniAD} & \multicolumn{1}{c}{1.26} & \multicolumn{1}{c}{0.29} &
    \multicolumn{1}{c}{1.24} & \multicolumn{1}{c}{0.66} \\
    \rowcolor[gray]{0.9} \textbf{Expert-UniAD} & \multicolumn{1}{c}{\textbf{1.24}} &
    \multicolumn{1}{c}{\textbf{0.24}} & \multicolumn{1}{c}{\textbf{1.08}} &
    \multicolumn{1}{c}{\textbf{0.57}}\\
    \midrule
    \textbf{VAD} & \multicolumn{1}{c}{1.41} & \multicolumn{1}{c}{0.38} &
    \multicolumn{1}{c}{1.38} & \multicolumn{1}{c}{0.75} \\
    \rowcolor[gray]{0.9} \textbf{Expert-VAD} & \multicolumn{1}{c}{\textbf{1.38}} &
    \multicolumn{1}{c}{\textbf{0.36}} & \multicolumn{1}{c}{\textbf{1.33}} &
    \multicolumn{1}{c}{\textbf{0.69}}\\
    \bottomrule
    \end{tabular}
    \end{adjustbox}
    \caption{\textbf{New-city generalization results.} We train the model on Boston and test it on Singapore, and vice versa, to evaluate the generalization capabilities.}
    \label{tab:generalization evaluation}
\end{table}

\subsection{Generalization Evaluation}\label{subsec:robustness_evaluation}
To assess generalization, we conduct cross-city tests on nuScenes by training on one city (Boston / Singapore) and testing on the other. As shown in Table~\ref{tab:generalization evaluation}, Expert-UniAD reduces collision rates by 17\% when trained on Boston and tested on Singapore. Conversely, when trained on Singapore, it achieves 13\% lower L2 errors and 14\% fewer collisions on Boston. Expert-VAD shows similar improvements across both settings. While Boston poses more challenging road conditions (higher open-loop metrics), models trained on Singapore still achieve greater gains on Boston, highlighting the strong generalization of \tool.

\subsection{Qualitative Study}\label{subsec:qualitative_results}
Fig.~\ref{Perception results} illustrates the effectiveness of \tool in a complex scenario. Here, UniAD fails to detect a traffic officer positioned ahead on the right, resulting in a planned path that veers toward the officer. In contrast, Expert-UniAD accurately detects the officer and immediately adjusts the route within the current frame, ensuring a safe trajectory. 

\section{Conclusions}
\label{sec:conclusion}

We have proposed \tool to improve the planning effectiveness and inference efficiency of ADSs with a Mixture of Experts architecture. Our evaluation has demonstrated the effectiveness and efficiency of \tool, and the contribution of each module in \tool. Further, \tool also exhibits strong generalization capability. Qualitative results and case studies further highlight its capabilities. 

\section{Acknowledgments}
This work was supported by the National Natural Science Foundation of China (Grant No. 92582205).

\appendix

\nobibliography*

\bibliography{aaai2026}

\end{document}